# Detecting Irregular Patterns in IoT Streaming Data for Fall Detection

Sazia Mahfuz
School of Computing
Queen's University
Kingston, Canada
mahfuz@cs.queensu.ca

Haruna Isah
School of Computing
Queen's University
Kingston, Canada
isah@cs.queensu.ca

Farhana Zulkernine
School of Computing
Queen's University
Kingston, Canada
farhana@cs.queensu.ca

Peter Nicholls
STSM
IBM Streams
IBM Canada
nicholls@ca.ibm.com

*Abstract*— Detecting patterns in real time streaming data has been an interesting and challenging data analytics problem. With the proliferation of a variety of sensor devices, real-time analytics of data from the Internet of Things (IoT) to learn regular and irregular patterns has become an important machine learning problem to enable predictive analytics for automated notification and decision support. In this work, we address the problem of learning an irregular human activity pattern, fall, from streaming IoT data from wearable sensors. We present a deep neural network model for detecting fall based on accelerometer data giving 98.75 percent accuracy using an online physical activity monitoring dataset called "MobiAct", which was published by Vavoulas et al. The initial model was developed using IBM Watson studio and then later transferred and deployed on IBM Cloud with the streaming analytics service supported by IBM Streams for monitoring real-time IoT data. We also present the systems architecture of the real-time fall detection framework that we intend to use with Mbientlab's wearable health monitoring sensors for real time patient monitoring at retirement homes or rehabilitation clinics.

## I. INTRODUCTION

In this digital era of Internet of Things (IoT) where hundreds of connected devices generate a massive amount of data every second, we face a dire challenge of storing and processing the data to generate usable knowledge and to create a smarter world. Detecting regularities and irregularities or anomalies in streaming data has the potential to provide insights and is useful in healthcare, finance, security, social media, and many applications [1]. Among the very many applications use cases reported in the literature are Global Positioning System (GPS) for locations, accelerometers, light and temperature sensors, gyroscopes for movements and orientations [2], system failure in manufacturing or industrial system [3], fall detection in human daily activities [4], intrusion detection [5], fraud detection in any financial system [6], and irregular reading detection in patient healthcare monitoring system. As a result, huge research interests and efforts are being geared towards the domain of pattern detection for real time streaming data in the IoT domain.

The notable research in this regard are the studies on detecting anomalies in real time streaming data [7-10] and outliers [11, 12]. Our aim is to explore the scenario of irregular human activity detection such as fall using real time sensor data. We map the problem as an irregular pattern detection problem, considering a fall as an irregular pattern in the context of regular human activity and attempt to distinguish the different fall scenarios from regular activities such as walking and sitting.

An estimated 646,000 fatal falls occur each year, making it the second leading cause of death because of unintentional injury, after road traffic injuries [13]. In all regions of the world, death rates are highest among adults over the age of 60 years. More than 50% of injury-related hospitalizations are seen in people aged over 65. Consequently, almost 40% of the injury related deaths are from falls in the elderly population [14]. In Canada many retirement and long-term care homes have very high patient to nurse ratios, and falls do not get reported until after some time. Falls also cause hip fracture, another common problem in the elderly population. For such reasons, fall detection and prevention has become an important problem to be solved.

In this paper, we present our ongoing work on building a framework for detecting irregular activity pattern of fall in IoT streaming data from wearable sensors used for monitoring human health and activity. Our work focuses on learning this irregular activity pattern from two annotated published datasets, MobiAct by Vavoulas et al. in 2016 [15], and SisFall by Sucerquia et al. in 2017 [16]. We develop and validate an artificial neural network (ANN) model for fall detection giving 98.75% accuracy. Next, we integrate the model into IBM Streams to build an IBM Cloud-based IoT data processing framework using real-time streaming sensor data. The research explores existing IBM tools to build a real-time machine learning framework to process and learn patterns in a variety of streaming IoT data.

The rest of the paper is organized as follows. Section 2 describes some background concepts relevant to our study. Section 3 describes the related work. The architecture of the system and the neural network model are described in Section 4. Section 5 presents and discusses the results. Section 6 discusses the conclusion and future work.

## II. Background

Data has the inherent characteristic of changing over time, and therefore, recognizing patterns in real time data involves dealing with several challenges including frequency, dimensionality, regularity, irregularity, outliers, and noise apart from the 4V challenges of big data namely volume, velocity, variety and veracity [17]. Furthermore, depending on the system status, regular patterns change with time and need to be redefined. In this research, we use streaming data from sensors that monitor human activity. Some concepts regarding human activity and streaming data relevant to our research are explained below.

### A. Streaming Data Proccessing Systems

Streaming data represents a continuous flow of data that needs to be processed by systems equipped to ingest, process, store and analyze the data. Existing data stream processing systems (DSPS) [18-21] typically include the following main components: (i) streaming data sources, (ii) data ingestion systems, (iii) data stream processing engines (DSPE), (iv) storage systems, (v) resource management services, and (vi) data sink to channel the output to other DSPSs, storage or visualization tools. Fig. 1 shows a general DSPS structure.

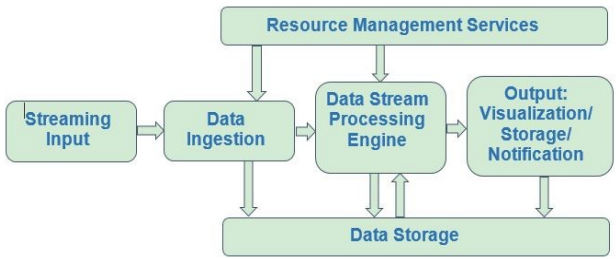

Fig. 1. A general structure for Data Stream Processing System (DSPS)

There are many open source and proprietary tools that exist in the market today which can be used to build the components of the DSPS represented in Fig.1. Typical examples of these cutting-edge tools include Kafka Connect, a framework included in Apache Kafka [22] for data stream ingestion, IBM Streams for data stream processing, Cassandra [23] for efficient storage, IBM Streams runtime system [24] for the management and administration of applications in the Streams runtime system, and Jupyter Notebook [25] for visualization. It is an ongoing challenge to study and compare the efficiency, suitability and compatibility of the new tools that are released every day with the existing technology and select the right tools to build a new DSPS or upgrade existing systems.

### B. Concept Drift

Data keep changing with time and situation. Therefore, the definition of regular and irregular patterns need to be updated as well. Streaming data generated from a particular source may not have the same statistical properties over time, which is referred to as concept drift [26]. An effective irregularity detection system should have the ability to handle concept drifts of streaming data as well as detect irregular patterns in a changing data. For example, after the onset of diabetes, a patient's normal blood sugar reading would be different compared to the reading before the diagnosis of diabetes. The normal blood sugar will remain different depending on the patient's specific disposition. Thus, the concept of the regular level of blood sugar must be updated in the model that detects irregular blood sugar level to accommodate such a concept drift. An effective real time irregularity detection system should be able to handle concept drift as it classifies irregularities in streaming data.

### C. Human Activities

Human activities include standard activities like walking, sleeping, lying, standing and all other activities like playing, cooking, running, falling. Fall is not a regular activity considered in an average human's daily routine. Fall detection or even better fall prediction will enable a healthcare provider to take better actions in those scenarios. Ozdemir and Barshan have identified 20 different fall scenarios and 16 activities of daily living (ADL) in their research work [27]. Kerdegari et al. have used 9 ADLs and 11 different fall activities [28]. Nukala et al. have used 7 ADLs, 7 dynamic gait indices and 10 different fall activities [29]. Similarly, other researchers have used their own versions of fall and ADL activities for carrying out the evaluation of the fall detection systems they had developed. There is no standard framework for defining ADLs and fall activities, which is a critical limitation of the research done in this domain. This is also recognized in the analysis done by Casilari-Perez et al. [14]. Our contributions on detecting irregular patterns in streaming data can serve as a benchmark for future researchers.

## III. Related Work

Our work proposes a deep learning model to detect fall and thereby, proposes a framework to use the model with online data from wearable sensor devices. As such we briefly discuss some recent related work on fall detection using IoT data and DSPS.

### A. Fall Detection

Khan and Hoey reviewed the fall detection techniques and held the opinion that fall should be considered as an abnormal activity and research should be progressed in that direction [4]. They have identified that most of the research work is done using simulated fall data because of the scarcity of real fall data as fall is a rare event. They have also suggested potential solutions using auto-encoder or recurrent neural network (RNN) for fall detection with few fall training data. As such, we are focusing our review on the systems that includes artificial neural networks (ANN) as one of the evaluated techniques.

Generally, Artificial Neural Networks (ANN) have consistently achieved better results in detecting fall from physical activity monitoring data. Ozdemir and Barshan have used 2,520 trials to create a large dataset that included 14 volunteers performing a standardized set of movements including 20 voluntary falls and 16 ADLs [27]. Their fall detection system achieved 95% accuracy by using multi-layer perceptron (MLP) for binary classification between ADL and fall. Kerdegari et al. recorded 1,000 movement acceleration data collected from 50 volunteers for 9 ADL activities and 11 different fall activities using a waist-worn accelerometer and obtained 91.6% accuracy for binary classification for ADL against fall using MLP [28]. Nukala et al. collected data from 322 tests done by volunteers performing 10 different falls, 7 ADLs and 7 dynamic gait index (DGI) tests and achieved 98.7% accuracy with MLP using scaled conjugate graduate learning [29].

Theodoridis et al. [30] developed two LSTM models, one with simple accelerometer data and another with accelerometer data rotated at an angle, using a published dataset called UR Fall Detection. The models were compared against a threshold-based approach and a support vector machine model for the same data and the comparative study showed that the LSTM models achieved better results. The LSTM model with rotation obtained the best results with 98.57% accuracy and 100% precision. Abbate et al. [31] developed a mechanism using a smartphone and a wearable sensor placed at the waist with a sampling frequency of 50Hz. When a fall is detected using the threshold-based technique it is sent to a classification engine for further analysis. The mechanism also has a notification option using which the user can turn off the false alarms. The false positive data is sent to the classification model again for training. A two-layer feed-forward model is used in the classification engine where the extracted features from the input signals are fed into the neural network model. Though the model achieved 100% accuracy on its training dataset from 7 volunteers, it would be interesting to check how the model performs with a larger dataset involving a higher number of older adults.

The work by Musci et al. used the SisFall dataset implementing Recurrent Neural Network (RNN) with underlying Long Short Term Memory (LSTM) blocks for developing online fall detection system [32]. They achieved 97.16% accuracy using an optimal window width of w = 256 sample with a custom weighted loss function after a thorough hyper-parameter optimization. Ajerla et al. [33] slightly modified the preprocessing done by Vavoulas et al. on the MobiAct dataset and achieved more than 90% accuracies on most of their experiments using MLP and LSTM [15, 33]. They have also achieved 99% accuracy using LSTM in two of their experiments. TABLE I compares the performance of the ANN techniques of the related work.

TABLE I. COMPARISON OF THE ANN TECHNIQUES

| Research Paper | Technique | Accuracy |
|---|---|---|
| Ozdemir et al., 2014 [7] | MLP | 95% |
| Kerdegari at al., 2013 [6] | MLP | 91.6% |
| Nukala et al., 2014 [8] | MLP | 98.7% |
| Theodoridis et al., 2018 [17] | LSTM | 98.57% |
| Abbate et al., 2012 [18] | Feed-forward ANN | 100% |
| Musci et al., 2018 [12] | RNN with LSTM | 97.16% |
| Ajerla et al., 2018 [11] | MLP and LSTM | 99% (LSTM) |

*B. DSPS*

According to Psaltis and de Assuncao et al. [19, 20] the DSPS architecture is generally multi-tiered and composed of many loosely coupled components comprising data sources, data collection, messaging systems, stream processing and delivery, and data storage systems.

Meehan et al. [18] developed a simple Extract, Transform, and Load (ETL) transactional DSPS architecture for processing IoT streams. The architecture comprised three main layers (i) stream collection for data ingestion, (ii) streaming ETL engine for real-time query processing, and (iii) an Online Analytical Processing (OLAP) backend for handling long-running queries. The connection between the ETL and OLAP components was provided by a data migration system.

The DSPS at Facebook [21] powers many use cases such as the real-time reporting of the aggregated voice of Facebook users, analytics for mobile applications, and insights for Facebook page administrators. It is made up of data sources such as mobile and web products; Scribe as a data distribution tool; stream processing systems such as Puma, Stylus, and Swift; and data stores such as Laser, Scuba, and Hive. The Facebook DSPS flow is a complex directed acyclic graph (DAG) comprising data streams from the mobile and web products that are fed to the Scribe. The real-time processing systems read and write data from and to Scribe. The data stores also use Scribe for data ingestion and serve different types of queries.

IV. PROPOSED SYSTEM ARCHITECTURE

*A. Datasets*

In the review on public datasets for wearable fall detection systems, Casilari-Perez et al. [14] have identified that the selection criteria for the dataset should prioritize the number of samples, experimental subjects, and the types of ADLs and falls included in the study. Based on this, we have selected two datasets, MobiAct and SisFall, for the evaluation of our system based on the number of subjects and number of activities covered by each dataset. MobiAct is chosen for the highest number of subjects whereas SisFall is chosen for the highest number of activities and inclusion of elderly people in their subject list. Table II summarizes the characteristics of the two datasets.

TABLE II. DATASETS USED FOR TRAINING THE MODEL

| Dataset | No. of subjects (female/male) | Age Range (no. of subjects) | No. of activities for ADLs /falls | No. of samples (ADLs /falls) |
|---|---|---|---|---|
| MobiAct [2] | 57 (15/42) | 20 - 47 | 9/4 | 2526 (1879/647) |
| SisFall [13] | 38 (19/19) | 19-30 (23) 60-75 (15) | 19/15 | 4505 (2707/1798) |

MobiAct by Vavoulas et al. (2016) [15] included labeled information for 4 different types of falls and 9 different ADLs collected from 57 subjects and more than 2,500 trials using a smartphone. The activities are represented using time stamp, raw accelerometer values, raw gyroscope values, and orientation data. Table III explains the activities covered in the MobiAct dataset. The limitation of the MobiAct dataset is that it does not include fall data from any elderly people. SisFall [16] remedies this scenario by including data from 15 elderly people aged between 60 and 75 years. Still, these two datasets do not include any real fall data.

SisFall by Sucerquia et al. [16] included annotated information for 15 different types of fall and 19 different types of ADLs collected from 38 subjects and more than 4,500 trials using a custom measuring device containing two different models of 3D accelerometers and a gyroscope positioned on a belt buckle.

TABLE III. ACTIVITIES COVERED IN MOBIACT DATASET

| | Code | Activity | Description |
|---|---|---|---|
| FALL | FOL | Forward-lying | Fall forward from standing, use of hands to dampen fall |
| | FKL | Front-knees-lying | Fall forward from standing, first impact on knees |
| | SDL | Sideward-lying | Fall sideward from standing, bending legs |
| | BSC | Back-sitting-chair | Fall backward while trying to sit on a chair |
| ADL | STD | Standing | Standing with subtle movements |
| | WAL | Walking | Normal walking |
| | JOG | Jogging | Jogging |
| | JUM | Jumping | Continuous jumping |
| | STU | Stairs up | 10 stairs up |
| | STN | Stairs down | 10 stairs down |
| | SCH | Sit chair | Sitting on a chair |
| | CSI | Car step in | Step in a car |
| | CSO | Car step out | Step out of a car |

## B. Data Preprocessing

*a) MobiAct:* We started our experiments with the complete set of features used by Ajerla et al. due to the high accuracy achieved by their LSTM model [33] and followed the same pre-processing steps. The raw data had a sampling rate of 20Hz and 10s duration. So every 200 records were aggregated to make one record for classifying the activity, and the most frequent class was used as the label for the concerned block.

Up to 58 features were generated using several statistical properties. For each axis (x, y, z) of the acceleration, 21 features were calculated from the mean, median, standard deviation (SD), skew, kurtosis, minimum and maximum. Using the absolute values of each axis (x, y, z) of the acceleration, another 21 features were calculated from the mean, median, SD, skew, kurtosis, minimum, and maximum. Another feature, slope, was calculated using Eq. 1, one for the given axes values (x, y, z) and another for the absolute values (|x|, |y|, |z|).

$$\text{Slope} = \sqrt{(max_x - min_x)^2 + (max_y - min_y)^2 + (max_z - min_z)^2} \quad (1)$$

Four other features were calculated using mean, SD, skew, and kurtosis of the tilt angle ($TA_i$) between the gravitational vector and the y-axis using Eq. 2.

$$TA_i = \sin^{-1}(y_i / (\sqrt{x_i^2 + y_i^2 + z_i^2})) \quad (2)$$

where $i$ represents the sample sequence.

Using the magnitude of the acceleration vector, 6 features were calculated from the mean, SD, minimum, maximum, difference between maximum and minimum, and zero crossing rate. Magnitude was calculated using Eq. 3.

$$Magnitude = \sqrt{x_i^2 + y_i^2 + z_i^2} \quad (3)$$

where $i$ represents the sample sequence.

For each axis (x, y, z) of the acceleration, the average absolute difference was calculated to obtain 3 more features. Also the average resultant acceleration of all the three axes was generated using Eq. 4.

$$Average\ resultant\ acceleration = \left(\frac{1}{n}\right) * \sum_i \sqrt{x_i^2 + y_i^2 + z_i^2} \quad (4)$$

where $i$ represents the sample sequence and n represents the total number of samples.

After feature extraction, the feature values were normalized using the min-max scaling formula given in Eq. 5.

$$X = \frac{X - Min(feature)}{Max(feature) - Min(feature)} \quad (5)$$

*b) SisFall:* Musci et al. did further annotation to the SisFall dataset as the dataset did not include timestamp data and Musci et al. required timestamp to implement LSTM [32]. But for our purpose, existing annotation in the SisFall dataset will suffice as we do not require timestamp data to implement our deep learning model.

Musci et al. used 14 characteristics to extract features from the dataset [16]. Among the extracted features, best results were observed for the characteristics coded $C_2, C_3, C_8, C_9$ and $C_{13}$ as given below. They showed the results using these characteristics in their paper. So, we are going to use the above-mentioned characteristics to extract features from the SisFall dataset as well. The metrics are given in Eq. 6 – 10.

$$C_2[k] = \sqrt{(a_x)^2[k] + (a_z)^2[k]} \quad (6)$$

$$C_3[k] = RMS(\max(\tilde{a}[k]) - \min(\tilde{a}[k])) \quad (7)$$

$$C_8[k] = \sqrt{(\sigma_x)^2[k] + (\sigma_z)^2[k]}; with\ \sigma_i = std(\tilde{a}_i[k]) \quad (8)$$

$$C_9[k] = \sqrt{(\sigma_x)^2[k] + (\sigma_y)^2[k] + (\sigma_z)^2[k]} \quad (9)$$

$$C_{13}[k] = \int \sqrt{(\tilde{a}_x)^2[n] + (\tilde{a}_z)^2[n]}\ dn \quad (10)$$

where *RMS* represents Root Mean Square value, $\sigma(.)$ represents the standard deviation operator, and $\tilde{a}[k]$ is defined in Eq. 11.

$$\tilde{a}[k] = \vec{a}^T[k - N_v + 1], ..., \vec{a}^T[k]; with\ \vec{a} = [a_x, a_y, a_z]^T \quad (11)$$

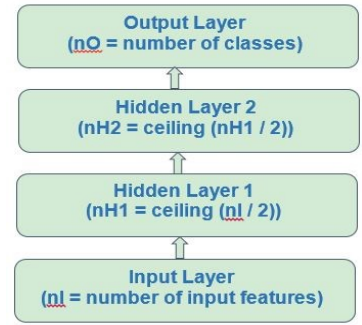

Fig. 2. Deep learning model used for the proposed system

## C. Predictive Modelling

We developed a simple Deep Neural Network (DNN) predictive model consisting of four layers including two hidden layers for detecting fall based on sensor data. The use of DNN is motivated by the successful use of MLP by other researchers

[17, 27-29, 33]. The DNN structure is shown in Fig. 2. The model was designed and built using offline datasets and later integrated and deployed in our streaming data processing framework as described below.

### D. Streaming Data Processing Framework

After testing and validating our predictive model with static fall data, we develop a streaming data processing framework with IBM tools. While this work is still ongoing, we present the system diagram and the implementation of the framework in this section. The system diagram is shown in Fig 3.

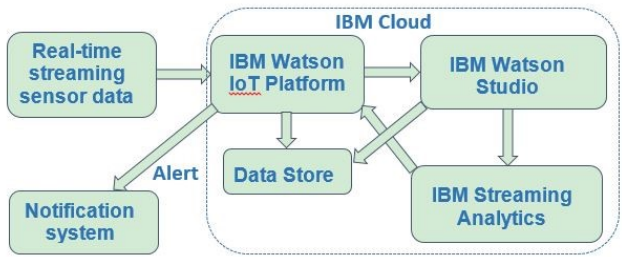

Fig. 3. The framework of the DSPS to detect irregular patterns for IoT streaming data

*1) Architecture and Components:* The proposed system has the following components.
- **Data Source:** Sensor data is retrieved from the smartphone positioned in the patient body.
- **Data Ingestion System:** We used IBM Watson IoT Platform to receive the sensor data from a smartphone or a wearable sensor and act as the Message Queuing Telemetry Transport (MQTT) [34] message broker. Apache Kafka [22] is used here as an open source alternative, which uses its own protocol.
- **Data Stream Processing Engine:** For offline data processing, we used IBM Watson Studio Notebook [35] as shown in Fig. 4, to run the feature extraction and to implement the machine learning model using the sample data. The Python code used in IBM Watson Studio Notebook and the intermediary data files have been posted on Github[1].

A key challenge is to train machine learning models using online data because long running processes can choke the data flow in stream processing engines. This part requires further research on IBM Streams, in-memory storage structures, and other open source tools such as Spark to build a multi-level in-memory data analytics framework. Such extensions to IBM Streams will enable extraction of streaming data to the in-memory storage structure and application of Spark in-memory analytics to train predictive models. We plan to use the IBM Streaming Analytics [36], a service for IBM Streams on IBM cloud, to continuously monitor the sensor data from smartphone or wearable sensors and send a notification to the monitoring application to alert healthcare providers about the emergency care needed by patients in the event o f a fall.

- **Data Store:** We plan to use IBM Cloudant [37] as the NoSQL database to store the sensor data from the Watson IoT platform and also the classification data obtained from the Watson Studio.
- **Data Sink:** The system will have multiple output channels.
  o Knowledge base: The regular and irregular patterns must be stored in a knowledge base to enable the creation of a patient profile.
  o Visualization: Visualization tools will be used to display the patterns.
  o Monitor: A signal can be generated when the irregular pattern is detected to notify appropriate authority to call for assistance. A tracker in the smartphone or an associated sensor data can be used to compute the patient's location to ensure that the patient gets the help quickly.
  o IBM Cloud: The message broker, data store, machine learning and the streaming processing system can be deployed on the IBM cloud for scalability with edge computing to reduce data transfer to the cloud.

*2) Implementation:* The experiments were run on a system with the following minimum hardware and software requirements:

Platform: Keras with TensorFlow backend
OS: Windows 10 64 bit
RAM: 8GB
CPU: Intel Core i-3-3217U, 1.80GHz
Hard Disk: 465 GB

To implement the system, we used a free IBM cloud account and the relevant services. We then added the following resources.
- Watson Studio
- Cloud Object Storage
- Streaming Analytics
- Internet of Things Platform

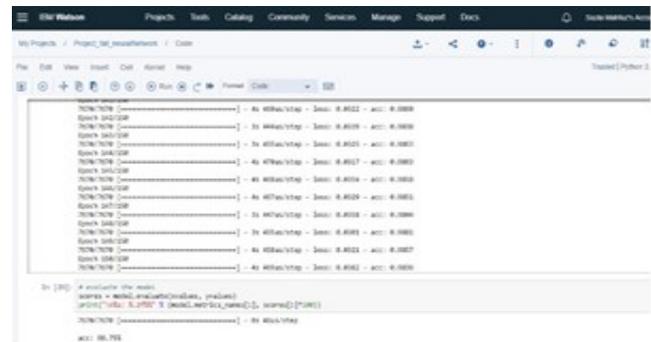

Fig. 4. Machine learning model executed using IBM Notebook

---

[1] https://github.com/SaziaM/CASCON2018

## E. Partial Results and Discussion

The machine learning model was first tested on the complete feature set for MobiAct as described in Section 4.2.1. This resulted in an accuracy of 98.75% for binary classification, fall and ADL (non-fall). The model was run for 150 epochs. A screenshot is shown for the IBM Watson Notebook execution in Fig 4.

The normalized and un-normalized confusion matrices for classifying fall and non-fall (ADL) cases out of 7,670 samples are shown in Fig. 5 and 6 respectively. This shows that the developed model performs quite well and can be used to develop the proposed system.

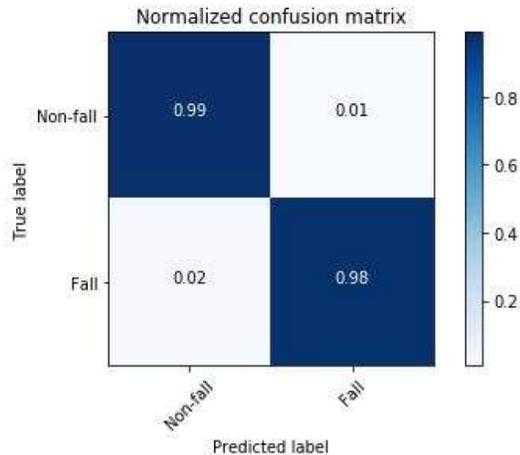

Fig. 5. Normalized confusion matrix for the trained model

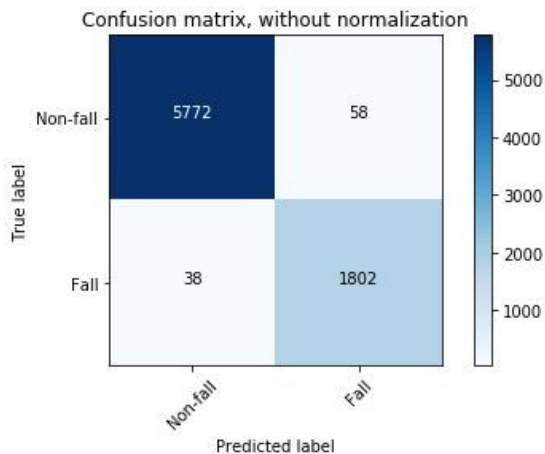

Fig. 6. Confusion matrix for the model without normalization

## V. CONCLUSION

Irregularity detection is an ongoing and critical data analytics problem for real time streaming data. In this paper, we have addressed this problem for the scenario of detecting irregularity, namely fall, in human activity data. Our proposed system will use real time physical activity monitoring data using wearable sensors, detect fall as an irregular pattern from the data, and notify the monitoring system for proper action in the event of a fall. This can be effectively used in healthcare centers or nursing homes to monitor elderly patients. We implemented a machine learning model for the proposed system using MLP which achieved 98.75% accuracy for the binary classification between fall and non-fall using the publicly available dataset called MobiAct.

For future work, we will extend the model to classify different types of fall, implement a real-time fall detection system for healthcare units using edge computing and IBM cloud. We plan to compare our system with other systems that use open source streaming data analytics tools to evaluate the functionality and performance of the IBM tools used in our framework. We also like to test the scalability and performance of our framework in connecting to and ingesting data from a large number of IoT devices.

## VI. ACKNOWLEDGEMENT

We like to thank IBM Canada, Queen's Centre for Advanced Computing (CAC), and Southern Ontario Smart Computing Innovation Platform (SOSCIP) for their research support in terms of funding, expert advice and access to high performance computing resources.